\documentclass[10pt,twocolumn,letterpaper]{article}

\usepackage[pagenumbers]{cvpr} 
\usepackage{graphicx}
\usepackage{subcaption}
\usepackage{booktabs}
\usepackage{graphicx}
\usepackage{multirow}
\usepackage{mathtools}
\usepackage{array}

\usepackage{kotex}
\usepackage{cuted}
\usepackage{currfile}
\usepackage{caption}

\preCutedStrip{\vspace{-5em}}

\definecolor{cvprblue}{rgb}{0.21,0.49,0.74}
\usepackage[pagebackref,breaklinks,colorlinks,allcolors=cvprblue]{hyperref}
\usepackage{algorithm}
\usepackage{algpseudocode}
\usepackage[most]{tcolorbox}
\usepackage{xcolor}

\definecolor{mygreen}{HTML}{147A4A}
\definecolor{myyellow}{HTML}{A07800}
\definecolor{myred}{HTML}{B4312B}
\definecolor{myorange}{HTML}{C46A1A}
\newtcolorbox{promptbox}[2][]{%
    enhanced,
    breakable,
    colback=gray!4,
    colframe=black,
    colbacktitle=black,
    coltitle=white,
    fonttitle=\bfseries\small,
    fontupper=\small,
    title={#2},
    boxrule=0.6pt,
    arc=1.2mm,
    outer arc=1.2mm,
    left=1.5mm,
    right=1.5mm,
    top=1.2mm,
    bottom=1.2mm,
    before skip=6pt,
    after skip=6pt,
    #1
}

\def\confName{CVPR}
\def\confYear{2027}

\iftoggle{cvprfinal}{%
  \title{PhysTrace: Interpretable Temporal Evaluation of Physical Consistency \\in Video Generative Models}
}{
    \title{PhysTrace: History-Conditioned Expectation Tracking for Physical Consistency \\in Video Generative Models}
}%

\author{
\begin{tabular}{@{\hspace{4pt}}c@{\hspace{8pt}}c@{\hspace{8pt}}c@{\hspace{8pt}}c@{\hspace{4pt}}}
Taekoan Yoo & 
Wonkyung Jung & 
Kyunghun Kim &
Kyeongbo Kong\textsuperscript{\textdagger} \\
\multicolumn{3}{c}{Independent Researcher, Korea} &
Pusan National University, Korea \\
{\tt\small tk.yoo@g.skku.edu} &
\multicolumn{2}{c}{{\tt\small \{gonyakjelly94,godroma4885\}@gmail.com}} &
{\tt\small kbkong@pusan.ac.kr}
\end{tabular}
}

\begin{document}
\maketitle
\begin{strip}
\vspace{12pt}
\centering

\includegraphics[width=0.99\textwidth]{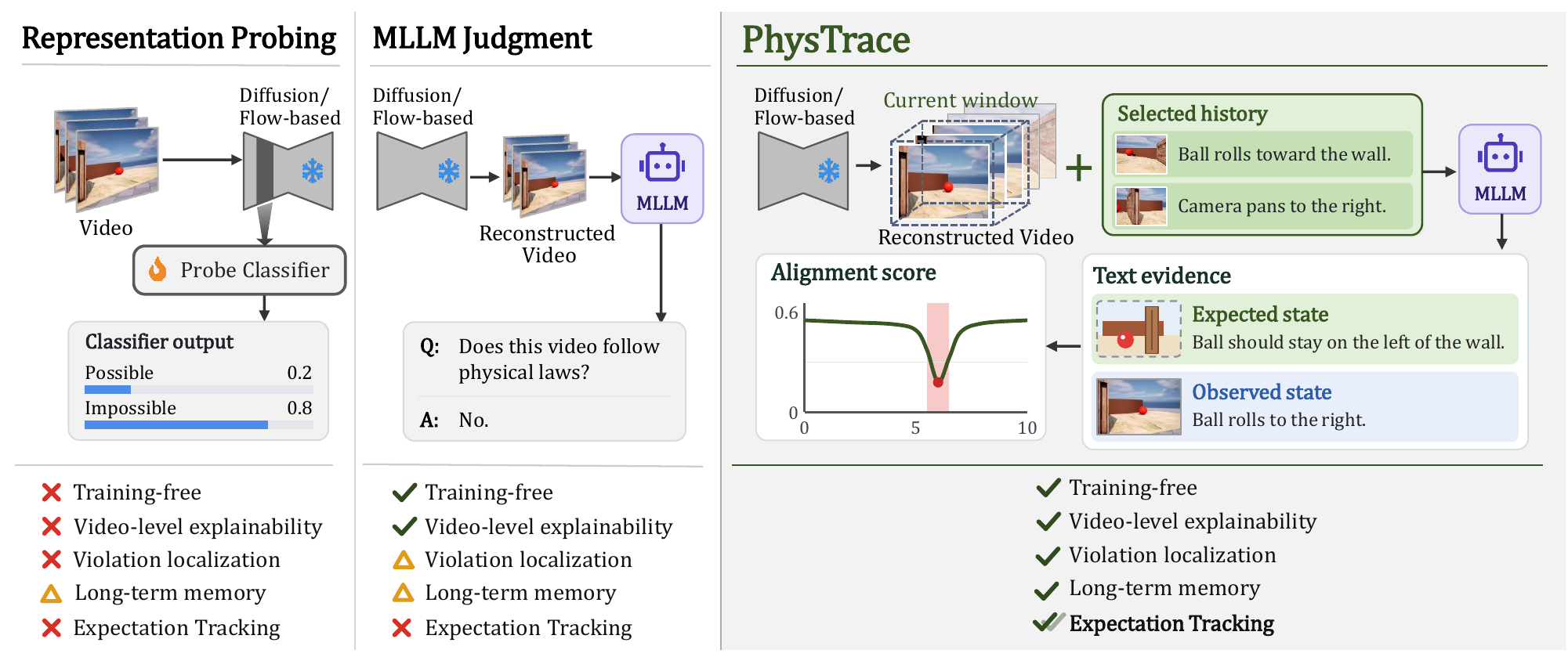}
\captionsetup{hypcap=false}
\captionof{figure}{
\textbf{Comparison of physical consistency evaluation approaches for diffusion/flow-based video generative models.}
Representation probing uses trained classifiers, while MLLM judgment provides direct video-level assessments.
PhysTrace instead tracks history-conditioned expectation--observation agreement to assess physical consistency without additional training, while also supporting temporal localization and interpretable diagnostics.
}


\captionsetup{hypcap=true}
\vspace{-10pt}
\label{fig:1}
\end{strip}

\begin{abstract}
World action models increasingly rely on pretrained diffusion/flow-based video generators assumed to capture physical dynamics for prediction, planning, and action generation.
Existing evaluations assess plausibility, probe representations, or measure predictive surprise, yet physical consistency is history-dependent: preceding events constrain what can physically occur later.
We introduce \emph{PhysTrace},\iftoggle{cvprfinal}{\footnote{\footnotesize\url{https://github.com/PhysTrace/PhysTrace}}}{} a training-free framework that uses MLLMs to formulate history-conditioned physical expectations and tracks their agreement with observed video windows through temporal video--language alignment, rather than relying on direct plausibility judgments or task-specific probes.
On IntPhys2, PhysTrace achieves 67.2\% possible--impossible discrimination accuracy, outperforming the evaluated direct MLLM baselines.
We further introduce reference-guided masked denoising for controlled reconstruction from mixed possible--impossible video latents.
To support these reconstructions and evaluate temporal localization as an additional capability, we introduce \emph{IntPhys2-Timing}, adding violation intervals to IntPhys2, with PhysTrace achieving a Hit Rate of up to 77.4\%.
On these reconstructions, human evaluations reveal model-dependent differences in physical consistency and fewer physically impossible outcomes at higher initial noise levels, trends broadly captured by PhysTrace.
\iftoggle{cvprfinal}{%
Videos illustrating our results are available at \\{\small\url{https://phystrace.github.io/PhysTrace.io/}}.
}{}
\end{abstract}
    
\section{Introduction}
\label{sec:1}

In robotic control, Vision-Language-Action (VLA) models have rapidly advanced by leveraging semantic knowledge and generalization capabilities acquired through large-scale vision-language pretraining for robot action generation~\citep{[1],[2],[3],[4],[5]}. More recently, research has increasingly explored leveraging future dynamics and predictive and spatiotemporal priors acquired through video pretraining for action prediction and planning~\citep{[6],[7],[8],[9],[10],[11]}. In particular, diffusion/flow-based generative video models have emerged as key pretrained backbones for these approaches, as they can directly generate future visual dynamics~\citep{[25],[26],[27],[28]}. This growing reliance on pretrained video models makes it increasingly important to determine whether they have learned the physical dynamics required for robotic control.

To this end, various benchmarks and evaluation methods have been proposed to assess intuitive physics, physical plausibility, and motion consistency in input or generated videos~\citep{[31],[32],[35],[36],[37],[38],[46]}. In parallel, studies have investigated whether diffusion/flow-based video generators encode physics-related information in their intermediate denoising representations~\citep{[41],[44],[45]}. 
As illustrated in Fig.~\ref{fig:1}, representation-probing approaches use trained classifiers on frozen model features to predict physical plausibility or other physical attributes.
Although such probes reveal whether physics-related information is recoverable from model representations, they do not directly establish whether generated dynamics remain physically consistent over time.

This distinction matters because physical consistency is not determined by the plausibility of individual frames alone, but by whether changes in object identity, state, motion, and interaction remain compatible with preceding observations. Earlier events constrain what can physically occur later, so an observation that looks plausible in isolation may become inconsistent when placed in its temporal context. Sparse or short observations may therefore miss these dependencies, while even a correct video-level judgment does not explicitly quantify whether subsequent observations remain consistent with physical expectations formed from preceding events.

A natural alternative to supervised probing is to directly ask a Multimodal Large Language Model (MLLM) to assess physical plausibility and explain its judgment.
However, \emph{Pixels to Principles}~\citep{[48]} reports near-chance performance of recent MLLMs in challenging intuitive-physics settings on GRASP~\citep{[49]} and IntPhys 2~\citep{[32]}, even though its probing analysis finds that relevant physical cues remain recoverable from visual representations in some settings. 
Regardless of judgment accuracy, a video-level plausibility assessment alone does not explicitly quantify how states expected from prior observations agree with subsequent observations over time.
Instead, we use the MLLM to formulate history-conditioned expected states and evaluate their agreement with the corresponding video evidence, rather than relying on its direct judgment as the consistency measure.

From this perspective, we propose \textit{PhysTrace}, a training-free framework that traces physical consistency in videos through history-conditioned temporal video-language alignment. 
As illustrated in Fig.~\ref{fig:2}, PhysTrace selectively retains salient past observations and uses an MLLM to formulate an Expected State for each current window under the accumulated temporal context. 
It aligns this expectation with the observed video window in a shared multimodal embedding space, producing a temporal alignment trace that serves as a diagnostic signal of physical consistency.
Drops in alignment additionally support violation localization, while the Expected State and Observed State descriptions provide interpretable evidence of discrepancies.
We evaluate the effectiveness of history-conditioned expectation tracking for physical consistency assessment through possible–impossible discrimination on IntPhys2~\citep{[32]}. 
We also introduce reference-guided masked denoising to investigate how pretrained video generators resolve mixed possible--impossible alternatives.
To support reconstruction and evaluate temporal localization, we construct \textit{IntPhys2-Timing} with annotated violation intervals and apply PhysTrace to the reconstructions to assess model-dependent physical consistency.

In summary, our contributions are threefold:

\begin{itemize}[leftmargin=*, labelsep=0.4em]
\vspace{-0.1em}
\setlength{\itemsep}{0.2em}
\setlength{\topsep}{0.1em}
\setlength{\partopsep}{0em}
\setlength{\parsep}{0em}

\item We propose \textit{PhysTrace}, a training-free framework that assesses physical consistency by aligning history-conditioned expectations with video observations over time, achieving 67.2\% possible--impossible discrimination accuracy on IntPhys2 and outperforming the evaluated direct MLLM and predictive-model baselines. Its alignment trace additionally supports interpretable violation localization.

\item We construct and release \textit{IntPhys2-Timing}, augmenting IntPhys2 with temporal violation annotations to support reference-guided reconstruction and temporal localization evaluation, where PhysTrace achieves a Hit Rate of up to 77.4\%.

\item We introduce reference-guided masked denoising to examine how pretrained video generators resolve mixed possible--impossible alternatives and apply PhysTrace to assess the physical consistency of the reconstructions. Human evaluations show fewer physically impossible outcomes at higher initial noise levels, a trend broadly reflected by PhysTrace.

\end{itemize}
\section{Related works}
\label{sec:2}

\subsection{VLA and world-model-based policies}
\label{sec:2.1}

VLA models couple pretrained vision-language representations with robot action generation, allowing large-scale multimodal knowledge to inform language-conditioned control~\citep{[1],[2],[3],[4],[5]}. Beyond this paradigm, recent work has explored leveraging predictive dynamics and spatiotemporal priors learned from video for robot policy learning, including predictive visual representations extracted from video diffusion models~\citep{[6]}.

More direct approaches, including World Action Models (WAMs) and related world-model-based policies, couple future prediction with action generation or planning~\citep{[7],[8],[9],[10],[11]}. Among these, DreamZero~\citep{[8]} jointly models future world states and actions using a pretrained video diffusion backbone, while Cosmos Policy~\citep{[10]} generates actions as latent frames and predicts future states and values for model-based planning. Despite their architectural differences, these approaches share the goal of exploiting predictive dynamics learned from video as priors for robot control.

\subsection{Pretrained video world models as policy priors}
\label{sec:2.2}

Prior work has explored different classes of video models for learning representations and dynamics relevant to robot control. Self-supervised video representation models learn temporal structure without explicit robot actions through latent prediction, as in V-JEPA2~\citep{[11]}, or masked video reconstruction~\citep{[22]}. A separate line of work develops robot-specific, action-conditioned world models. DreamDojo~\citep{[23]}, for example, learns interaction dynamics from large-scale human videos and is subsequently adapted to generate future visual states conditioned on robot actions.

Another line of work builds general-purpose generative video models, such as Stable Video Diffusion~\citep{[25]}, Wan~\citep{[26]}, and the Cosmos-Predict family~\citep{[27],[28],[29]}. 
These diffusion/flow-based models directly generate temporally evolving visual sequences and have increasingly been repurposed as pretrained backbones for representation-based policies and WAMs~\citep{[6],[7],[8],[10]}. Unlike latent predictive or robot-specific world models, our work focuses on this class of pretrained generative video models and examines whether their generated dynamics preserve physical consistency. 
Among the models evaluated, Wan2.1 provides the pretrained video backbone adopted by DreamZero~\citep{[8]}, while Cosmos-Predict2 is adapted for visuomotor control and planning in Cosmos Policy~\citep{[10]}.

\subsection{Physical understanding and probing in video models}
\label{sec:2.3}

A broad range of benchmarks evaluate physical understanding in videos, including input-video benchmarks for intuitive physics and physical reasoning~\citep{[31],[32],[33],[34]} and evaluations of physical plausibility and motion consistency in generated videos~\citep{[35],[36],[37],[38]}. 
Among them, IntPhys2~\citep{[32]} evaluates four intuitive-physics principles---permanence, immutability, spatio-temporal continuity, and solidity---through video-level discrimination between physically possible and impossible events, but lacks temporal violation annotations.
While PhyScore~\citep{[64]} provides temporal anomaly annotations for world-model-generated videos, such temporal annotations are unavailable for the controlled intuitive-physics scenarios.
We therefore introduce \textit{IntPhys2-Timing}, which augments impossible videos with annotated violation intervals.

In parallel, probing and interpretability studies investigate physics-related information in pretrained video representations~\citep{[39],[40]}, comparing pretraining paradigms~\citep{[41]}, decoding physical plausibility from intermediate denoising representations~\citep{[44],[45]}, and evaluating model-native preferences through the denoising objective~\citep{[46]}.

Beyond whole-video scoring, TRAJAN~\citep{[38]} localizes motion inconsistencies, while SPOTLIGHT~\citep{[65]} uses direct VLM judgments to localize and explain generated-video errors, including physics violations, without explicitly constructing reference expectations or quantifying their discrepancies with observations.
VoE-inspired predictive approaches~\citep{[80],[81],[32]} measure time-resolved surprise by comparing predicted future representations with observations, but their timing can depend on the prediction horizon, and their scores do not explicitly identify conflicting history-conditioned physical expectations.
PhysTrace instead retains salient past states and aligns history-conditioned expectations with the same current window to quantify expectation--observation agreement for physical consistency assessment, while additionally supporting localization and interpretable diagnostics (Fig.~\ref{fig:1}).


\section{PhysTrace: temporal physical consistency tracing}
\label{sec:3}
\begin{figure*}[t]
    \centering
    \includegraphics[width=0.90\textwidth]{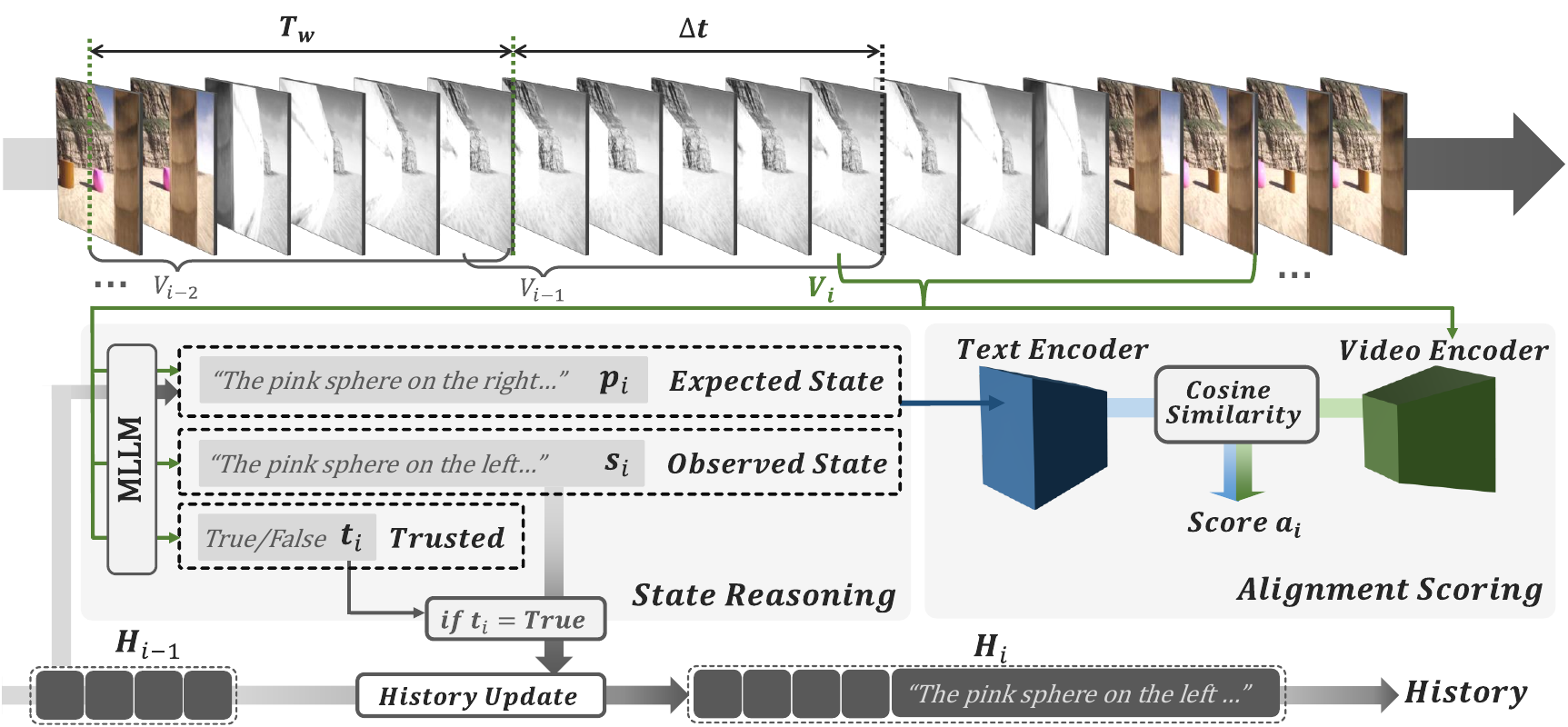}
    \caption{
    \textbf{Overview of PhysTrace.}
        History-conditioned expected states are aligned with current video windows, while salient observations are selectively retained in history. Grayed-out frames indicate observations excluded from history ($t_i=0$). Alignment uses either a unified multimodal encoder or separate contrastively aligned video and text encoders.
    }
    \vspace{-10pt}
    \label{fig:2}
\end{figure*}

In this section, we introduce PhysTrace, which traces changes in physical consistency over time by repeatedly measuring the alignment between expected video dynamics and actual observations. 
We also describe the evaluation metrics for video-level physical consistency and temporal violation localization, together with IntPhys2-Timing.
The overall framework is illustrated in Fig.~\ref{fig:2}.

\subsection{Temporal expectation--observation alignment}
\label{sec:3.1}

Given a video $V$, we divide it into temporal windows of fixed duration $T_w$ with a hop size of $\Delta t$. For each window $V_i$, a fixed number of frames are sampled and provided to an MLLM using attribute-specific prompts with a common structured output format.

At each window, the MLLM generates two textual descriptions: the \texttt{Observed State} $s_i$, describing the objects and dynamics directly observed in $V_i$ without using history as visual evidence, and the \texttt{Expected State} $p_i$, describing the state expected in the current window based on the accumulated history $\mathcal{H}_{i-1}$. 
Predictive VoE formulations~\citep{[80],[81],[32]} typically quantify surprise by comparing a predicted future representation with its subsequent observation, potentially shifting the detected discrepancy according to the prediction horizon. PhysTrace instead generates the history-conditioned \texttt{Expected State} $p_i$ for the current window and compares it directly with the same $V_i$\iftoggle{cvprfinal}{  (see Appendix~\ref{app:C}).}{  (see Appendix C).}

\begin{table}[t]
\centering

\caption{Cosine distance ($\downarrow$) between MLLM-generated descriptions and ground-truth captions using Qwen3-Embedding~\citep{[63]}. All Trusted: all \texttt{Trusted} indicators set to True; Guided: fixed history and \texttt{Trusted} indicators.}
\vspace{-10pt}
\label{tab:1}
\small
\setlength{\tabcolsep}{3pt}

\begin{tabular}{l|ccc|c}
\hline
\multirow{2}{*}{\textbf{Model}}& \multicolumn{3}{c|}{\textbf{Expected}}
& \textbf{Observed} \\
\cline{2-4} \cline{5-5}
& All Trusted & Ours & Guided & Guided \\
\hline

Qwen3-VL (32B)
& 0.229 & 0.204 & 0.156 & 0.149 \\
Qwen3.8 (27B)
& 0.281 & 0.270 & 0.191 & 0.188 \\
Cosmos-Reason2 (8B)
& 0.350 & 0.342 & 0.270 & 0.242 \\
GLM-4.1V
& 0.268 & 0.246 & 0.222 & 0.220 \\

\hline
Swapped Prompts
& \multicolumn{3}{c|}{0.500}
& 0.501 \\
\hline
Irrelevant Prompts
& \multicolumn{3}{c|}{0.669}
& 0.672 \\
\hline
\end{tabular}
\vspace{-20pt}
\end{table}

To examine whether MLLMs can reliably generate the \texttt{Observed State} and \texttt{Expected State} descriptions, we manually annotate reference descriptions for each temporal window of the original IntPhys2 videos~\citep{[32]} and evaluate several MLLMs in Table~\ref{tab:1}.
In the Guided setting, history and \texttt{Trusted} (Sec.~\ref{sec:3.2}) are fixed to isolate description generation. The Irrelevant Prompts and Swapped Prompts baselines measure cosine distances between ground-truth descriptions and unrelated text prompts or descriptions from unrelated videos, respectively. 
The generated \texttt{Observed State} and \texttt{Expected State} descriptions in the Guided setting achieve substantially lower cosine distances to ground truth than both baselines, supporting the ability of MLLMs to generate meaningful descriptions of current observations and history-conditioned expected states under controlled conditions. 
This supports their role in PhysTrace's contextual reasoning rather than direct judgments of physical validity.

Since the resulting \texttt{Expected State} $p_i$ is directly used for video--language alignment, we constrain it to a concise visual caption containing only visually verifiable information, such as object identity, visual attributes, and motion. An alignment scoring model then maps $V_i$ and $p_i$ into a shared embedding space, using the same sampled frames provided to the MLLM.
Let $f_{\mathrm{video}}$ and $f_{\mathrm{text}}$ denote the video and text embedding functions, implemented using either separate encoders or a unified multimodal embedding model\iftoggle{cvprfinal}{ (see Appendix~\ref{app:A}).}{ (see Appendix A).} The expected-state--observation alignment score $a_i$ for the $i$-th window is computed using cosine similarity as follows:
\vspace{-5pt}
\begin{equation}
a_i =
\frac{
f_{\mathrm{video}}(V_i)^{\top}
f_{\mathrm{text}}(p_i)
}{
\left\|f_{\mathrm{video}}(V_i)\right\|_2
\left\|f_{\mathrm{text}}(p_i)\right\|_2
}.
\label{eq:1}
\end{equation}

Repeating this computation across all temporal windows yields the temporal alignment trace, $\mathcal{A}(V) = \{a_1, a_2, \ldots, a_N\}$.

\begin{figure*}[t]
    \centering
    \includegraphics[width=0.99\textwidth]{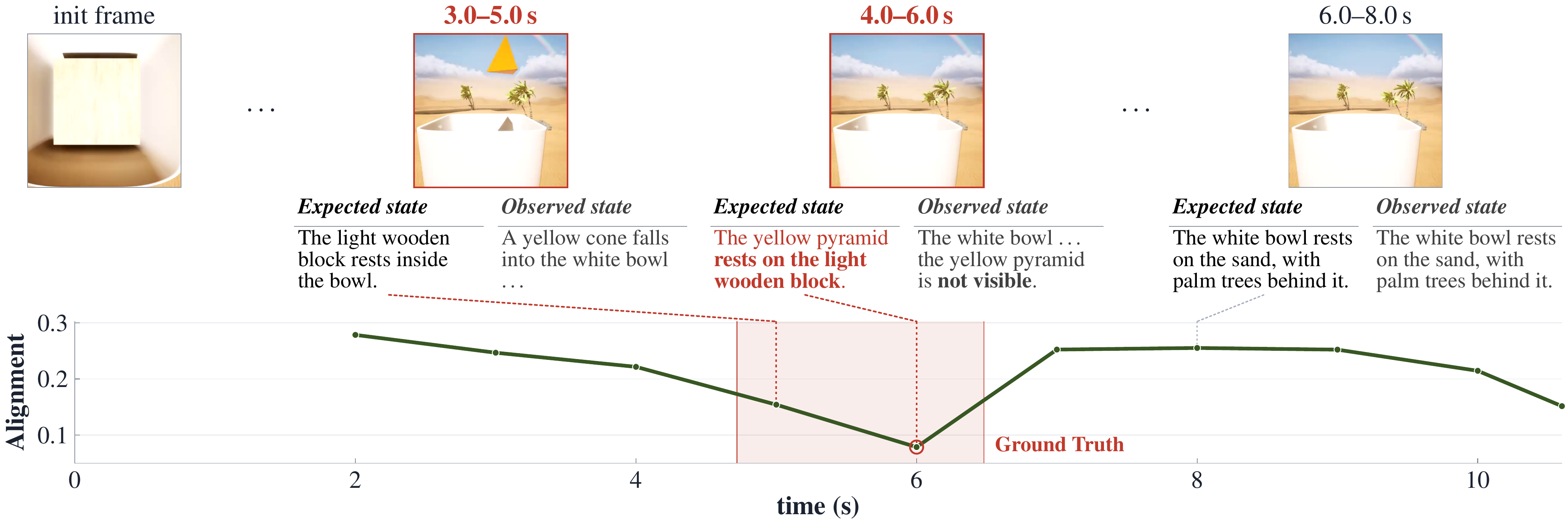}
    \vspace{-10pt}
    \caption{\textbf{Example of physical violation localization for solidity.} \textit{Observed State} describes the current window, while \textit{Expected State} describes how the window should appear given the trusted history on IntPhys2-Timing.}
    \vspace{-10pt}
    \label{fig:3}
\end{figure*}

When the state expected from previous observations is preserved in the current window, the alignment between the \texttt{Expected State} $p_i$ and observation $V_i$ tends to remain relatively high. 
Conversely, observations that conflict with history-conditioned expectations, such as unexplained object disappearance, abrupt changes in object identity, or physically implausible motion and collisions, tend to reduce alignment.
As illustrated in Fig.~\ref{fig:3},  PhysTrace uses these variations as a diagnostic signal of potential physical inconsistency and to identify candidate violation intervals.

To assess video-level physical consistency, we summarize the temporal alignment trace using the \textit{Minimum Alignment Score (MAS)}, defined as $\mathrm{MAS}(V)=\min_i a_i$. 
A lower MAS reflects a larger mismatch between history-conditioned expectations and observations and is used to compare possible--impossible video pairs.

\subsection{History-conditioned expectations and selective state retention}
\label{sec:3.2}

Physical violations can occur at different temporal scales. Short temporal windows capture brief, localized changes, whereas longer events or violations involving temporally separated events require broader temporal context. Simply increasing the window size, however, may obscure these short events and reduce temporal localization precision. PhysTrace therefore maintains local temporal windows while selectively retaining salient observations from preceding windows in the history $\mathcal{H}_{i-1}$.

In addition to \texttt{Observed State} and \texttt{Expected State} described in Sec.~\ref{sec:3.1}, the MLLM produces a \texttt{Trusted} indicator \(t_i \in \{0,1\}\) for each window to determine whether the currently observed state \(s_i\) is retained in the history for subsequent expected-state generation. To avoid accumulating repetitive or uninformative observations, \(t_i=1\) is assigned only when a new object appears, new motion begins, or a meaningful state change not already represented in the history occurs. The history is updated as follows:
\begin{equation}
\mathcal{H}_i =
\begin{cases}
\mathcal{H}_{i-1} \cup \{s_i\}, & \text{if } t_i = 1, \\
\mathcal{H}_{i-1}, & \text{if } t_i = 0.
\end{cases}
\label{eq}
\end{equation}

As shown in Table~\ref{tab:1}, selective history retention (Ours) consistently yields lower cosine distances for \texttt{Expected State} than retaining all observations (All Trusted). Distances decrease further in the Guided setting, where history and \texttt{Trusted} are fixed, demonstrating the importance of appropriate temporal context for generating history-conditioned expected states.

If the \texttt{Observed State} $s_i$ is retained in the history after a physical violation, subsequent expected states are conditioned on the updated history, allowing the alignment trace to emphasize the initial deviation from expected physical behavior rather than remaining low throughout subsequent windows. 
\iftoggle{cvprfinal}{Additional qualitative examples are provided in Appendix~\ref{app:B}.
}{}

\subsection{IntPhys2-Timing: annotations and evaluation}
\label{sec:3.3}

The original IntPhys2 provides only video-level possible--impossible annotations, without temporal violation locations. To support temporal localization evaluation and reference-guided reconstruction (Sec.~\ref{sec:4}), we construct \textit{IntPhys2-Timing}\iftoggle{cvprfinal}{\footnote{\footnotesize\url{https://github.com/PhysTrace/PhysTrace/tree/main/IntPhys2Timing}}}{} by augmenting the impossible videos in IntPhys2 with temporal violation annotations. Of the original 506 video pairs, we exclude seven invalid pairs with identical possible and impossible videos and annotate violation intervals for all 499 remaining impossible videos.
For each impossible video, we use its corresponding possible video as a reference to identify where their physical behaviors begin to diverge, then manually refine the boundaries to annotate the actual violation interval. 

For temporal localization evaluation, we retain eight of the 16 categories, excluding those with ambiguous object identities, substantial occlusion, cluttered scenes, or atypical structures that complicate reliable temporal localization evaluation.
The minimum-score location in each impossible video corresponds to a temporal interval of length $\Delta t$, the window hop size.
If this interval overlaps the annotated violation interval, the video is counted as a hit; \textit{Hit Rate} is the proportion of annotated impossible videos satisfying this criterion.





\section{Reference-guided reconstruction of physical alternatives}
\label{sec:4}

In this section, we examine pretrained diffusion/flow-based video generative models, including Wan2.1~\citep{[26]} and Cosmos-Predict2~\citep{[29]}, whose backbones have also been adopted for action models.
To investigate how these models resolve physically possible versus impossible alternatives, we introduce reference-guided masked denoising that allows mixed paired-video latents to evolve toward either outcome under adaptive guidance.

\subsection{Reference-guided masked denoising}
\label{sec:4.1}

For each video pair in the full IntPhys2-Timing dataset, we restrict generative changes to the annotated violation interval while preserving the shared pre-violation context.
The procedure consists of five steps.

\noindent\textbf{(i) Paired latent references.}
Given a possible--impossible video pair $(V_P,V_I)$, we obtain their VAE-encoded clean latents $z_P=E_{\mathrm{VAE}}(V_P)$ and $z_I=E_{\mathrm{VAE}}(V_I)$.
We define a binary mask $M$ over the ground-truth violation interval, using the masked midpoint $z_{\mathrm{mix}}=M\odot(z_P+z_I)/2$ inside the mask and the shared pre-violation latent $z_{\mathrm{pre}}=(1-M)\odot z_I$ outside it, preserving the common pre-violation context of both videos.

\noindent\textbf{(ii) Masked initialization and context reset.}
We start denoising at a fixed initial noise level
$\sigma_{\mathrm{init}}$.
Using the masked midpoint $z_{\mathrm{mix}}$ inside the violation interval and the unmasked reference latent $z_{\mathrm{pre}}$ outside it, we construct the initial noisy latent by adding Gaussian noise $\varepsilon\sim\mathcal{N}(0,I)$:
\begin{equation}
\begin{aligned}
z_{\sigma_{\mathrm{init}}}
={}&(1-\sigma_{\mathrm{init}})
(z_{\mathrm{mix}}+z_{\mathrm{pre}})
+\sigma_{\mathrm{init}}\varepsilon.
\end{aligned}
\label{eq:masked_init}
\end{equation}
Throughout denoising, the masked region follows the denoising updates, while the unmasked region is reset at every noise level $\sigma_j$ to $(1-\sigma_j)z_{\mathrm{pre}}+\sigma_j(1-M)\odot\varepsilon_j$, using a RePaint~\citep{[88]}-inspired context reset.
This ensures that only the masked region evolves through denoising while the shared pre-violation context remains anchored to its reference.

\noindent\textbf{(iii) Relative proximity scoring.}
At each denoising step $j$, the model predicts a clean latent estimate $\hat{z}_0^{(j)}=\mathcal{R}_{\theta}(z_{\sigma_j},\sigma_j)$.
We measure its squared distances to each reference $k\in\{P,I\}$ within the mask:
\begin{equation}
d_k^{(j)}
=\|M\odot(\hat{z}_0^{(j)}-z_k)\|_2^2.
\label{eq:latent_dist}
\end{equation}
The relative possible-reference score is
\begin{equation}
r_P^{(j)}=
\operatorname{sigmoid}\left(
\frac{d_I^{(j)}-d_P^{(j)}}{\tau_{\mathrm{rel}} d_{PI}}
\right),
\label{eq:latent_score}
\end{equation}
where $d_{PI}=\|M\odot(z_P-z_I)\|_2^2$ normalizes the distance difference by the squared separation between the two references, and $\tau_{\mathrm{rel}}=0.1$ controls the score sensitivity.
Equal distances yield $r_P^{(j)}=0.5$, while values approaching $1$ or $0$ indicate greater proximity to the possible or impossible reference, respectively.

\noindent\textbf{(iv) Adaptive latent guidance.}
We construct a reference-weighted target and guide the predicted clean latent toward it:
\begin{equation}
\begin{aligned}
z_{\mathrm{tar}}^{(j)}
&=r_P^{(j)}z_P+(1-r_P^{(j)})z_I,\\
\tilde{\hat{z}}_0^{(j)}
&=\hat{z}_0^{(j)}
+\lambda M\odot
(z_{\mathrm{tar}}^{(j)}-\hat{z}_0^{(j)}),
\end{aligned}
\label{eq:latent_guidance}
\end{equation}
where $\lambda$ controls the guidance strength.
Only the masked region is modified, and the guided clean estimate is converted to the model's native prediction parameterization before advancing the denoising scheduler.

\noindent\textbf{(v) Hard selection and reconstruction.}
When the noise level reaches a threshold $\sigma_{\mathrm{hard}}$, we binarize the relative possible-reference score defined in Eq.~\eqref{eq:latent_score}:
\begin{equation}
r_P^{(j)}\leftarrow
\mathbf{1}[r_P^{(j)}\geq 0.5],
\quad \sigma_j\leq\sigma_{\mathrm{hard}}.
\label{eq:hard_selection}
\end{equation}
This selects either $z_P$ or $z_I$ as the guidance target for the remaining steps.
After denoising, we restore the clean unmasked context and decode the final latent:
\begin{equation}
\hat{V}=D_{\mathrm{VAE}}\left(
M\odot z_0^{\mathrm{final}}
+z_{\mathrm{pre}}
\right).
\label{eq:masked_decoding}
\end{equation}


\subsection{Video reconstruction and evaluation setup}
\label{sec:4.2}

To apply PhysTrace to pretrained diffusion/flow-based video generative models, we account for their model-specific input configurations. For example, Wan2.1 and Cosmos-Predict2 process 81 and 93 frames per segment, respectively, at a nominal rate of 16 fps, whereas the original IntPhys2~\citep{[32]} videos contain 636 frames (60 fps, approximately 10.6 s) at a resolution of $512 \times 512$. 
Rather than subsampling frames for single-pass reconstruction, we process full-length IntPhys2-Timing video pairs through successive overlapping segments.
For each segment, we apply the paired-latent initialization and masked denoising procedure described in Sec.~\ref{sec:4.1}.
For PhysTrace evaluation, we use the same temporal windowing and frame-sampling procedure as for the original videos, providing identical sampled frames to both the MLLM and alignment scoring model. Finally, to avoid the influence of scene descriptions on the reconstructed outputs, we set the video models' text-conditioning prompts to an empty string in all generative-trajectory experiments.

\section{Experiments}
\label{sec:5}

For video--text alignment scoring, we use PE-AV-Large~\citep{[58]}\iftoggle{cvprfinal}{ (see Appendix~\ref{app:A})}{ (see Appendix A)}, with Qwen3.8-27B as the default MLLM unless otherwise specified.
We process the original IntPhys2 videos using temporal windows of $T_w = 2$s with a hop size of $\Delta t = 1$s, uniformly sampling 60 frames per window for both the MLLM and the alignment scoring model.
To generate history-conditioned expected states, we use separate attribute-specific MLLM prompts sharing a common structured output format.
Experiments requiring local inference are conducted on NVIDIA A100 80GB GPUs.

\subsection{Possible-vs-impossible classification accuracy on IntPhys2}
\label{sec:5.1}

Table~\ref{tab:2} evaluates possible--impossible discrimination on IntPhys2~\citep{[32]}.
For the MLLM and predictive-model baselines, we adopt the evaluation approaches used in IntPhys2: MLLMs directly classify each video as physically possible or impossible, while predictive models such as V-JEPA2~\citep{[11]} compare prediction-based surprise between paired videos.
In contrast, PhysTrace compares the MAS of each possible--impossible pair, counting the pair as correctly discriminated when the impossible video has a lower MAS than its possible counterpart.

All three PhysTrace variants improve over direct judgments by their corresponding MLLMs in overall accuracy.
While Pixels to Principles~\citep{[48]} reports near-chance performance of MLLMs in challenging intuitive-physics settings, our results suggest that tracing history-conditioned expectation--observation discrepancies provides a more effective signal for physical consistency discrimination than direct MLLM judgments.

PhysTrace also outperforms the compared predictive-model baselines. Unlike prediction-based surprise, which can depend on the prediction horizon, PhysTrace directly aligns explicitly formulated, history-conditioned expectations with observations in the same temporal window, suggesting a more effective approach to assessing physical consistency.

\subsection{Temporal localization of physical violations on IntPhys2-Timing}
\label{sec:5.2}

\begin{table}[t]
\centering
\caption{\textbf{Possible-vs-Impossible accuracy (\%) on the IntPhys2 benchmark across difficulty levels.} \textbf{Med.}: Medium; \textbf{Unk.}: Unknown. Results marked with $^*$ are taken from prior work.}
\vspace{-7pt}
\label{tab:2}

\small
\setlength{\tabcolsep}{4pt}

\begin{tabular}{l|c@{\hspace{3pt}}c@{\hspace{3pt}}c@{\hspace{3pt}}c|c}
\toprule
\textbf{Model} & \textbf{Easy} & \textbf{Med.} & \textbf{Hard} & \textbf{Unk.} & \textbf{Overall} \\
\midrule
Ours (Qwen3.8-27B) & \textbf{67.3} & {73.5} & \textbf{63.7} & \textbf{59.3} & \textbf{67.2} \\
Ours (Gemini-3.8-Flash) & {65.4} & \textbf{77.0} & {58.9} & 57.0 & {66.4} \\
Ours (GPT-5.4) & {55.8} & {68.0} & {61.9} & 59.3 & {63.2} \\
\midrule

Qwen-VL 2.5-72B& 51.9 & 49.3 & 50.0 & 48.3 & 49.6 \\
Qwen 3.8 (27B)& 58.7 & 56.8 & 48.5 & 52.3 & 53.5 \\
Gemini-2.5 (Pro)& {62.5} & 54.8 & 53.6 & 56.4 & 55.4 \\
Gemini-2.5(Flash) & 52.9 & 57.3 & 55.1 & 54.7 & 55.6 \\
Gemini-3.8(Flash) & {63.5} & 60.8 & 56.3 & 62.2 & 59.8 \\
GPT-4o & 51.0 & 54.0 & 51.2 & 51.2 & 52.3 \\
GPT-5.4 & 53.9 & 53.5 & 52.7 & 55.2 & 53.6 \\
\midrule
VideoMAEv2 & 42.3 & 52.0 & 48.2 & 51.2 & 49.6 \\
V-JEPA2-h & 42.3 & 50.5 & 50.6 & 50.0 & 49.6 \\
V-JEPA2-g & 34.6 & 55.5 & 41.7 & 57.0 & 49.0 \\
HERA (w/ RRPM)*~\citep{[82]} & - & - & - & - & 54.4 \\
PhiZero*~\citep{[83]} & 61.0 & 60.5 & 52.4 & - & 56.3 \\

\bottomrule
\end{tabular}
\vspace{-10pt}
\end{table}

\begin{table}[t]
\centering
\caption{\textbf{Temporal localization hit rates on IntPhys2-Timing.}
Overall Hit Rates (\%) aggregated across physical attributes and difficulty categories. \iftoggle{cvprfinal}{Detailed results are provided in Appendix~\ref{app:E}.}{Detailed results are provided in Appendix E.}}
\label{tab:3}
\vspace{-7pt}
\small
\setlength{\tabcolsep}{4pt}

\begin{tabular}{l|c}
\toprule
\textbf{Model} & \textbf{Overall} \\
\midrule
Ours (Qwen3.8-27B~\citep{[89]}) & 73.1 \\
Ours (Gemini-3.8-Flash~\citep{[90]}) & \textbf{77.4} \\
Ours (GPT-5.4~\citep{[91]}) & 70.3 \\
\midrule
VoE (V-JEPA2-h) & 44.9 \\
VoE (V-JEPA2-g) & 61.1 \\
\midrule
Window-based MLLM & 38.9 \\
Window-based MLLM (w/ history) & 60.4 \\
\bottomrule
\end{tabular}
\vspace{-5pt}
\end{table}

We additionally evaluate PhysTrace's temporal localization capability on IntPhys2-Timing, focusing on physical violations that depend on preceding events, using VoE-inspired and window-based MLLM baselines.
The VoE baselines~\citep{[80],[81],[32]}, implemented using V-JEPA2-h and V-JEPA2-g~\citep{[11]}, compute time-resolved prediction surprise using the same temporal context and target-window settings as PhysTrace, selecting the interval with the highest surprise.
Similarly, window-based MLLMs assign violation scores to temporal windows and select the highest-scoring interval, using the same window duration and hop size as PhysTrace. We evaluate two variants, with and without history, to examine the contribution of preceding observations.
A prediction is counted as a hit if its predicted interval overlaps the annotated violation interval.

As shown in Table~\ref{tab:3}, PhysTrace achieves overall Hit Rates of 70.3--77.4\%, outperforming both VoE baselines and window-based MLLMs.
In particular, incorporating history into window-based MLLMs improves the Hit Rate from 38.9\% to 60.4\%, highlighting the importance of temporal context for identifying physical violations.
These results further support the value of history-conditioned expectation tracking for temporally grounded physical consistency assessment.
\begin{table}[t]
\centering
\caption{\textbf{Ablation study of history and selective retention on IntPhys2.}
Possible-vs-impossible classification accuracy (\%) across difficulty levels.
\textbf{Med.}: Medium; \textbf{Unk.}: Unknown.}
\label{tab:4}
\vspace{-7pt}
\small
\setlength{\tabcolsep}{4pt}

\begin{tabular}{cc|cccc|c}
\toprule
\textbf{History} & \textbf{Selective}
& \textbf{Easy} & \textbf{Med.} & \textbf{Hard}
& \textbf{Unk.} & \textbf{Overall} \\
\midrule
 &  & 50.0 & 50.5 & 49.4 & 52.3 & 50.4 \\
$\checkmark$ &  & 65.4 & 71.0 & 60.7 & 57.0 & 64.6 \\
$\checkmark$ & $\checkmark$
& \textbf{67.3} & \textbf{73.5} & \textbf{63.7}
& \textbf{59.3} & \textbf{67.2} \\
\bottomrule
\end{tabular}
\vspace{-10pt}
\end{table}

\subsection{Ablation study on history}
\label{sec:5.3}

\begin{figure*}[t]
    \centering
    \includegraphics[width=0.90\textwidth]{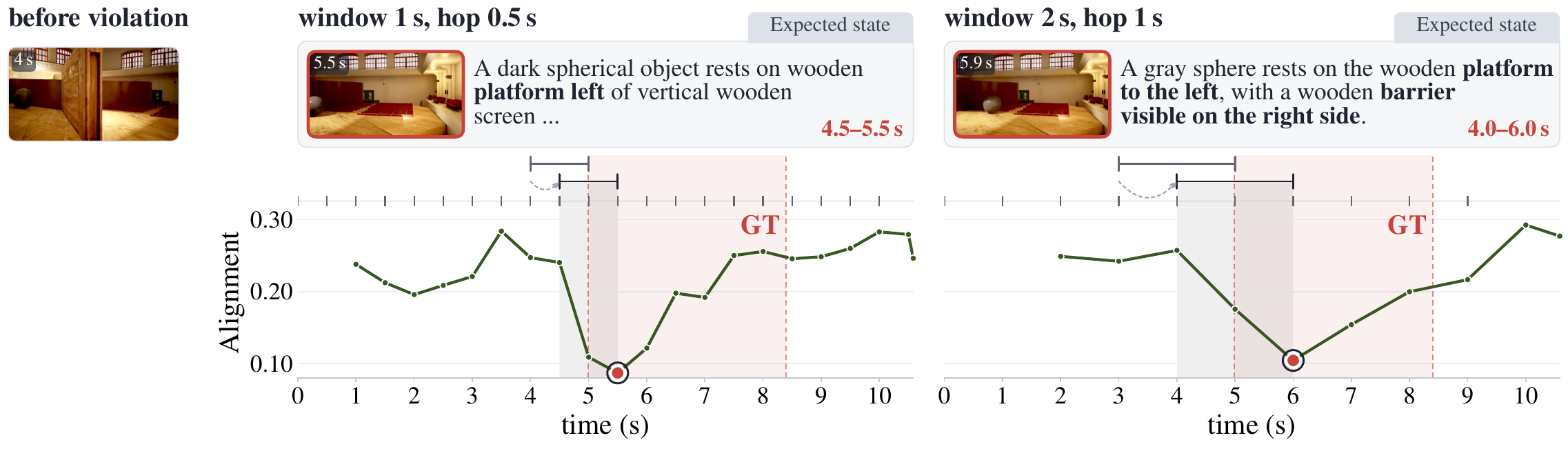}
    \caption{\textbf{Effects of window and hop sizes.} Broader temporal coverage makes the violation more salient, yielding a sharper drop within the ground truth (GT) interval for the solidity attribute on IntPhys2-Timing.}
    \vspace{-10pt}
    \label{fig:4}
\end{figure*}

\begin{table*}[t]
\centering
\small

\caption{\textbf{Physically impossible classification rates (\%, $\downarrow$) on reference-guided reconstructions from the full \textit{IntPhys2-Timing} dataset.}
Results are reported for human evaluators and PhysTrace across video models and initial noise levels. \textbf{Bold} and \underline{underlined} values indicate the lowest and highest rates, respectively, for each evaluator and noise level.}

\label{tab:5}

\setlength{\tabcolsep}{4pt}
\vspace{-5pt}
\begin{tabular}{l@{\hspace{0pt}}c@{\hspace{6pt}}c@{\hspace{3pt}}c@{\hspace{3pt}}c@{\hspace{10pt}}|>{\hspace{3.2em}}c@{\hspace{4pt}}c@{\hspace{4pt}}c|>{\hspace{3.2em}}c@{\hspace{4pt}}c@{\hspace{4pt}}c}
\toprule

\multirow[b]{2}{*}{\textbf{Model}}
& \multirow[b]{2}{*}{\textbf{Params}}
& \multirow[b]{2}{*}{\textbf{Backbone}}
& \multirow[b]{2}{*}{\textbf{Formulation}}
& \multirow[b]{2}{*}{\textbf{VAE}}
& \multicolumn{3}{c|}{\textbf{Human Evaluation}}
& \multicolumn{3}{c}{\textbf{PhysTrace}} \\
\cmidrule(lr){6-8}
\cmidrule(l){9-11}
& & & & &
$\mathllap{\sigma_{\mathrm{init}}=\,}0.5$ &
$0.8$ &
$1.0$ &
$\mathllap{\sigma_{\mathrm{init}}=\,}0.5$ &
$0.8$ &
$1.0$ \\
\midrule






ModelScope-1.7B~\citep{[84]}
& 1.7B & U-Net & DDPM & 2D
& \underline{70.3} & 53.7 & \textbf{47.4}
& \underline{62.5} & 59.5 & \textbf{56.3} \\

VideoCrafter2~\citep{[85]}
& 1.4B & U-Net & DDPM & 2D
& \textbf{60.9} & \textbf{51.3} & 48.3
& \textbf{59.5} & \textbf{58.3} & 57.3 \\

CogVideoX-5B~\citep{[86]}
& 5B & DiT & DDPM & 3D
& 62.4 & 54.9 & \underline{51.9}
& 61.1 & 60.1 & \underline{58.9} \\

Wan2.1-T2V-1.3B~\citep{[26]}
& 1.3B & DiT & Flow Matching & 3D
& 65.3 & \underline{57.2} & 48.0
& 62.3 & \underline{61.1} & 58.3 \\

LongCat-Video~\citep{[87]}
& 13.6B & DiT & Flow Matching & 3D
& 65.9 & 56.5 & 49.7
& 62.3 & 60.5 & 58.7 \\

Cosmos-Predict2-2B~\citep{[29]}
& 2B & DiT & EDM & 3D
& 63.1 & 56.3 & 50.5
& 61.3 & 59.7 & 58.1 \\

\bottomrule
\end{tabular}%

\vspace{-10pt}
\end{table*}

Table~\ref{tab:4} examines the contributions of history conditioning and selective state retention to physical consistency discrimination on IntPhys2.
Without history, accuracy remains near chance level, while incorporating history substantially improves performance.
Selective retention further improves accuracy across all difficulty levels, highlighting the importance of relevant temporal context for expectation tracking.

\subsection{Ablation study on temporal windowing}
\label{sec:5.4}

Fig.~\ref{fig:4} compares alignment traces for $(T_w,\Delta t)=(2s,1s)$ and $(1s,0.5s)$.
The longer-window setting yields a sharper alignment drop within the violation interval, whereas the shorter setting provides finer temporal resolution but exhibits greater score fluctuations.
We therefore adopt $T_w=2s$ and $\Delta t=1s$ to balance temporal context and localization stability.



\subsection{Physical tracing in reference-guided reconstructions}
\label{sec:5.5}





Table~\ref{tab:5} evaluates the physical consistency of reference-guided reconstructions generated by diffusion/flow-based video models using the procedure described in Sec.~\ref{sec:4.1}.
Using possible--impossible video pairs from the full IntPhys2-Timing dataset, the reconstruction process allows mixed latents to evolve toward either outcome.
We set $\sigma_{\mathrm{hard}}=0.01$ and $\lambda=0.5$ for the reported results.
We classify a reconstructed video as physically impossible when its MAS is lower than that of the corresponding original possible video.
We report impossible-classification rates from PhysTrace and three human evaluators who independently classified all reconstructions as possible or impossible.

Qualitative results reveal model-dependent tendencies toward physically possible or impossible outcomes\iftoggle{cvprfinal}{\footnote{\footnotesize\url{https://github.com/PhysTrace/PhysTrace}}.}{ (see supplementary videos).}
Human evaluation shows that reconstructions initialized closer to clean latents more frequently yield physically impossible outcomes, whereas higher initial noise levels lead to more balanced possible--impossible outcomes.
Differences across models also narrow as the initial noise level increases.
This pattern is consistent with longer denoising trajectories providing greater opportunity to resolve mixed physical alternatives, with the extent of this change varying across models.

PhysTrace likewise identifies fewer physically impossible reconstructions at higher initial noise levels and broadly reflects model-dependent trends observed in human evaluation, although its variations are less pronounced.
These results suggest that history-conditioned expectation tracking captures meaningful differences in the physical consistency of generated outcomes.
\iftoggle{cvprfinal}{Detailed model-wise trends across initial noise levels are provided in Appendix~\ref{app:reconstruction}.}{Detailed model-wise trends across initial noise levels are provided in Appendix F.}

\section{Conclusion}

We introduce \textit{PhysTrace}, a training-free framework that assesses physical consistency through history-conditioned expectation--observation alignment, together with \textit{IntPhys2-Timing}, which supports reference-guided reconstruction and temporal localization evaluation.
PhysTrace achieves 67.2\% overall possible--impossible discrimination accuracy on IntPhys2, outperforming the evaluated direct MLLM and predictive-model baselines, while also supporting temporal violation localization.
Applying PhysTrace to reference-guided reconstructions further reveals model-dependent differences in physical outcomes.
Human evaluations show fewer physically impossible reconstructions at higher initial noise levels, a trend also reflected by PhysTrace.
Future work will focus on improving possible--impossible discrimination accuracy to enable more reliable comparisons of physical consistency across video generative models.

\clearpage
\newpage
\section*{LLM Usage Statement}

OpenAI's ChatGPT was used to assist with English translation, grammatical correction, editorial refinement, and citation preparation.
Anthropic's Claude assisted with code refactoring, iterative experiment execution, and revisions to selected figures.
Claude was also used to develop the entire project demo webpage.
The pretrained generative models used as components of the proposed method and experimental evaluations are described in the corresponding sections of the paper.
The authors reviewed the AI-assisted content and remain responsible for all scientific claims, experimental results, and conclusions.

{
    \small
    \bibliographystyle{ieeenat_fullname}
    \bibliography{main}
}

\iftoggle{cvprfinal}{%
  \clearpage
\setcounter{page}{1}
\maketitlesupplementary
\appendix

\providecommand{\mainref}[2]{%
    \ifdefined\standaloneSupplement
        #2%
    \else
        \ref{#1}%
    \fi
}

\section{Alignment model selection and temporal window configuration}
\label{app:A}

This section describes the criteria used to select the alignment model and discusses how its frame-sampling strategy affects the temporal-window configuration in PhysTrace.

To measure alignment between video and text, we consider models that map the two modalities into a shared embedding space. These include video--text representation models with separate visual and text encoders trained using cross-modal contrastive objectives~\citep{[51],[53],[54],[55]}, as well as multimodal retrieval embedders such as PE-AV~\citep{[58]}, which learns unified representations across multiple modalities. When selecting an alignment model, retrieval performance alone is insufficient; its frame-sampling strategy must also be considered because it directly determines how much temporal information can be retained within each window.

For example, in our PhysTrace setting, InternVideo2-CLIP~\citep{[54]} represents each video clip, corresponding to one temporal window, using eight sampled frames, whereas PE-AV-Large~\citep{[58]} supports variable-length video inputs and can use all frames within a window. As described in Sec.~\mainref{sec:3.1}{3.1}, a video $V$ is divided into temporal windows of length $T_w$, which are analyzed while being shifted by a hop size $\Delta t$. A larger $T_w$ provides broader temporal context and can capture events that unfold over longer durations, but sufficiently dense frame sampling is required to preserve short or transient events within each window.

The IntPhys2~\citep{[32]} videos used in our benchmark are typically recorded at 60 fps and contain approximately 636 frames ($10.6$ s). With InternVideo2-CLIP~\citep{[54]}, representing such a window with only eight sampled frames can therefore result in sparse temporal coverage as $T_w$ increases, potentially missing brief but important events. In contrast, PE-AV-Large~\citep{[58]} can exploit substantially denser frame information within the same temporal window. 
Beyond its strong alignment performance discussed below, these sampling characteristics provided an additional motivation for selecting PE-AV-Large as the alignment model for computing the alignment score.

To evaluate the alignment performance of the candidate models, we focus on their sensitivity to small caption errors, since the alignment model in PhysTrace scores candidate captions against each temporal window. Table~\ref{tab:app1} evaluates this sensitivity using 24 WebVid clips~\citep{[76]} and 86 perturbed captions, each created by replacing a caption element related to the subject, location, action, or attribute. $Det.\ Rate$ denotes the percentage of perturbations that reduce the video--text alignment score relative to the original caption, while $\Delta Det.$ measures the normalized score drop, where -100\% corresponds to the drop induced by replacing the caption with an unrelated sentence. Fig.~\ref{fig:app1} further illustrates how $\Delta Det.$ varies across perturbation types for each model.

While $Det.\ Rate$ is comparable across the evaluated models, PE-AV-Large achieves the largest magnitude of $\Delta Det.$ ($-18.9\%$), approximately $1.6\times$ that of InternVideo2-Stage2-6B ($-12.1\%$). Thus, PE-AV-Large detects caption errors at a similar frequency while penalizing detected mismatches more strongly, further supporting its use as the alignment model in PhysTrace.


\begin{table}[t]
\centering
\caption{\textbf{Caption substitution sensitivity on WebVid.} \textbf{$Det.\ Rate$}: fraction of caption-element substitutions that lower the original alignment score. $\Delta Det.$: mean signed score change normalized by the mean drop from unrelated-sentence replacements ($-100\%$ = as large as the unrelated-sentence drop). }

\label{tab:app1}

\small
\setlength{\tabcolsep}{4pt}
\renewcommand{\arraystretch}{1.1}

\begin{tabular}{lcc}
\toprule
\textbf{Alignment Scoring Model}
& \textbf{$Rate$\ (\%)}
& $\Delta$ \textbf{(\%)} \\
\midrule

PE-AV-Large~\citep{[58]}
& 84.9 & \textbf{-18.9} \\

PE-Core-G14-448~\citep{[59]}
& \textbf{88.4} & -15.3 \\

PE-Core-L14-336~\citep{[59]}
& 86.0 & -15.2 \\

InternVideo2-CLIP-6B~\citep{[54]}
& \textbf{88.4} & -17.5 \\

InternVideo2-Stage2-6B~\citep{[54]}
& 79.1 & -12.1 \\

VidVec embedding (Qwen3-VL-8B)~\citep{[60]}
& 79.1 & -13.5 \\

\bottomrule
\end{tabular}
\vspace{-15pt}
\end{table}

\section{Qualitative physical consistency traces}
\label{app:B}
Fig.~\ref{fig:app2} presents qualitative examples of PhysTrace applied to IntPhys2-Timing across four physical principles. The MLLM uses general contextual reasoning over object attributes such as shape, size, color, and structure, as well as spatial position, motion, and interactions with other objects, to formulate history-conditioned expected states.
By aligning these expectations with actual video observations, PhysTrace traces history-conditioned physical consistency over time, with the corresponding descriptions providing qualitative evidence of expectation–observation discrepancies.

The four physical principles in IntPhys2~\citep{[32]} characterize different aspects of object behavior. 
Object permanence requires objects to persist even when occluded, with unexpected disappearance or appearance indicating a violation. 
Object immutability requires objects to preserve their intrinsic properties, such as shape, color, and structure, without abrupt transformations. 
Spatio-temporal continuity requires objects to move continuously through space and time, without sudden changes in position or discontinuous motion. 
Finally, solidity requires solid objects to occupy distinct spaces, so passing through a solid obstacle or reacting to a nonexistent obstacle constitutes a physical violation.

\paragraph{Potential threshold-based violation detection.}
The temporal alignment trace could also support video-level violation detection by applying a threshold to the Minimum Alignment Score (MAS).
For example, with $\tau_{\mathrm{det}}=0.16$, videos with $\mathrm{MAS}(V)>\tau_{\mathrm{det}}$ could be treated as having no detected violation, while the minimum-score interval could be retained as a candidate violation otherwise.
Under such a criterion, an impossible video with no detected violation would be counted as a miss in temporal localization, regardless of interval overlap.



\section{Prompt design and selection}
\label{app:C}

This section describes the prompt-design criteria and input configurations used in our experiments.
As described in Sec.~\mainref{sec:3.2}{3.2}, for each temporal window, the MLLM produces three outputs: \texttt{Observed State}, \texttt{Expected State}, and \texttt{Trusted}. Based on preliminary experiments, we established the following prompt-design criteria to obtain reliable outputs for each component.
\begin{itemize}
    \item \texttt{Observed State}, \texttt{Expected State}, and \texttt{Trusted}: All three outputs are instructed to focus primarily on the foreground objects rather than incidental background changes or camera motion. In particular, the prompts emphasize object persistence and visibility, intrinsic properties such as shape, size, color, and structure, spatial position and motion, and interactions with other objects. These attributes capture object-centric cues that are broadly relevant to reasoning about physical consistency over time.

    \item \texttt{Observed State},\texttt{Trusted}: The model must describe the current state using only the current short video clip, without allowing information from the history to alter what is observed in the current window.

    \item \texttt{Expected State}: The expected object state should be determined primarily from the trusted history. The current \texttt{Observed State} may be used to identify the current scene phase, motion, and position, but not to overwrite object properties established by the trusted history. If the current observation is consistent with the trusted history, \texttt{Expected State} should remain consistent with the \texttt{Observed State}.

    \item \texttt{Expected State}: The model should not produce an unconstrained expectation while ignoring the current state. In particular, \texttt{Expected State} describes what should be visible in the current window according to the trusted history, rather than blindly expecting a future clip.

    \item \texttt{Trusted}: Any information considered for retention must be grounded only in the current video clip. History may be used only to determine whether that information is already represented, not to alter or infer the content observed in the current window.

    \item \texttt{Trusted}: Information already represented in the history should be assigned \texttt{False}. Excessively assigning \texttt{True} causes the history to grow unnecessarily and reduces the relative importance of salient information; therefore, only novel and task-relevant information should be retained.

    \item \texttt{Trusted}: Conversely, information that is not yet represented in the history but is necessary for subsequent expectations should be assigned \texttt{True}. Excessively assigning \texttt{False} can leave insufficient information in the history to construct reliable expectations in later windows.
\end{itemize}

In addition to these design criteria, the prompts are instantiated with several input parameters that specify the demonstration setting and the temporal information associated with each window:
\begin{itemize}
    \item \texttt{num\_prompt} (system): the number of demonstration captions ($n$);
    \item \texttt{segment\_number} (user): the one-based index of the current window;
    \item \texttt{start\_sec} (user): the start time of the current window in seconds;
    \item \texttt{end\_sec} (user): the end time of the current window in seconds;
    \item \texttt{next\_overlap\_sec} (user): the temporal overlap between adjacent windows, equal to the window length minus the hop size;
    \item \texttt{next\_duration\_sec} (user): the duration of the next window, corresponding to the window length.
\end{itemize}

We additionally follow model-specific reasoning configurations for each MLLM.
For Cosmos-Reason2~\citep{[62]}, the explicit \texttt{<think>} instruction is appended to the user prompt rather than the system prompt.
Qwen3-VL-Thinking~\citep{[61]} and Qwen3.8-27B~\citep{[89]} natively support reasoning within \texttt{<think>...</think>} tags and require no such instruction.
Gemini-3.8-Flash~\citep{[90]} and GPT-5.4~\citep{[91]} instead handle reasoning through their respective API configurations without explicit reasoning-format instructions in the prompts.

\section{Output-space readout via video inversion}
\label{app:D}

As a complementary analysis of physical consistency along generative trajectories, we use the clean-video estimate predicted from the current latent state at each noise level as an output-space readout. We obtain latent trajectories from VAE-encoded reference video content $z_0$ through deterministic inversion~\citep{[44],[74]}, which follows the sampling dynamics in reverse from clean states toward noise.

At each trajectory checkpoint, we obtain a clean latent estimate $\hat{z}_0^{(t)}$ from the model's native prediction and reconstruct the corresponding video $\hat{x}^{(t)}$ using the VAE decoder. This process is expressed as
\begin{equation}
\hat{z}_0^{(t)}
=
\mathcal{R}_{\theta}(z_t,t),
\qquad
\hat{x}^{(t)}
=
D_{\mathrm{VAE}}\left(\hat{z}_0^{(t)}\right),
\label{eq:trajectory_readout}
\end{equation}
where $\mathcal{R}_{\theta}$ denotes a model-specific readout that estimates the clean latent from the current trajectory state. For Wan2.1~\citep{[26]}, $\hat{z}_0^{(t)}$ is computed from the predicted velocity, whereas Cosmos-Predict2~\citep{[29]} uses a clean-latent estimate obtained from its denoiser. Although their readout implementations differ, both provide decoded clean-video estimates $\hat{x}^{(t)}$ as inputs to PhysTrace.

\begin{figure}[t]
    \centering
    \includegraphics[width=1.0\columnwidth]{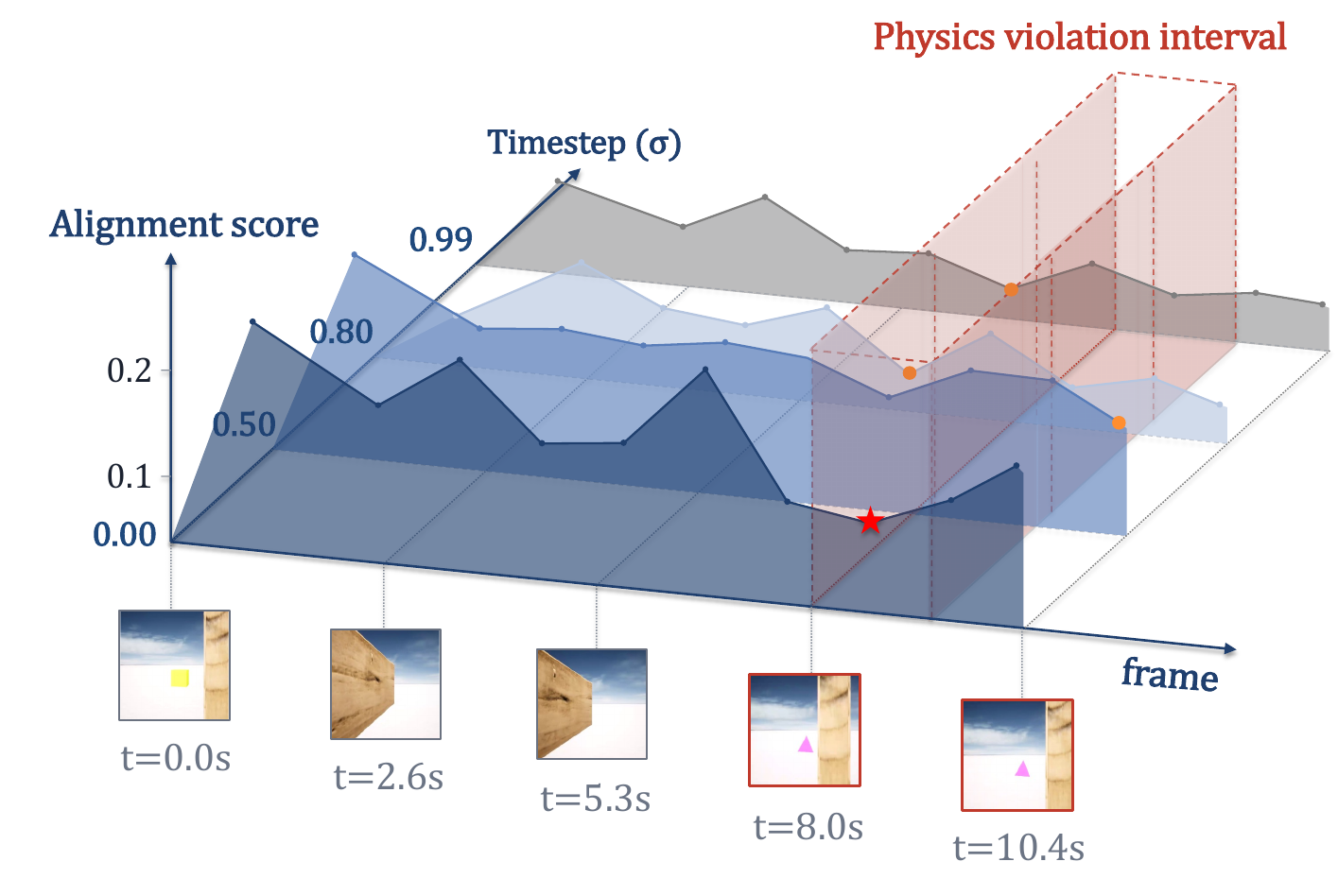}
    \vspace{-18pt}
    \caption{\textbf{Temporal alignment across diffusion timesteps.} Alignment traces for a spatio-temporal continuity example at different noise levels $\sigma$, computed with Wan2.1-T2V-1.3B. \textcolor{myred}{Red star}: minimum alignment score within the violation interval. \textcolor{myorange}{Orange dots}: minimum alignment scores outside the violation interval.}

    \label{fig:app0}
    \vspace{-10pt}
\end{figure}
We apply PhysTrace to $\hat{x}^{(t)}$ at multiple noise-level checkpoints, tracing how physics-related information is reflected in reconstructed video outputs rather than probing hidden representations directly~\citep{[44]}. As shown in Fig.~\ref{fig:app0}, high noise levels may leave scene structure and object dynamics insufficiently reconstructed, resulting in low alignment scores that provide limited evidence of physical consistency. As noise decreases, visual dynamics become clearer and the alignment trace reveals discrepancies between history-conditioned expectations and observations.


\section{Detailed temporal localization results}
\label{app:E}

Table~\ref{tab:app2} provides a detailed breakdown of the temporal localization results in Sec.~\mainref{sec:5.2}{5.2} across physical attributes and difficulty levels.
PhysTrace generally achieves higher Hit Rates than the compared VoE and window-based MLLM baselines, with particularly strong performance on immutability, although the improvements vary across attributes and difficulty categories.

\begin{table}[t]
\centering
\caption{\textbf{Temporal localization hit rates on IntPhys2-Timing.} Hit Rates (\%) across physical attributes and difficulty categories. VoE: a V-JEPA2-h-based surprise baseline. \textbf{Med.}: Medium; \textbf{Unk.}: Unknown. MLLM variants: {Qwen} = Qwen3.8-27B~\citep{[89]}; {Gemini} = Gemini-3.8-Flash~\citep{[90]}; {GPT} = GPT-5.4~\citep{[91]}.}
\label{tab:app2}
\vspace{-7pt}
\small
\setlength{\tabcolsep}{4pt}

\begin{tabular}{@{\hspace{0pt}}c@{\hspace{3pt}}l@{\hspace{3pt}}c@{\hspace{3pt}}c@{\hspace{3pt}}c@{\hspace{3pt}}c|c|c@{\hspace{0pt}}}
\toprule
\textbf{Method}
& \textbf{Attribute}
& \textbf{Easy}
& \textbf{Med.}
& \textbf{Hard}
& \textbf{Unk.}
& \textbf{All}
& \textbf{Avg}\\
\midrule




\multirow{4}{*}{\shortstack[c]{Ours\\[3pt](Qwen)}}
& Continuity & 75.0 & 83.3 & 69.2 & 50.0 & {72.4} & \multirow{4}{*}{\textbf{73.1}}  \\
& Immutability & 100.0 & 96.7 & 84.0 & 58.3 & {86.7} & \\
& Permanence & 50.0 & 66.7 & 61.5 & 58.3 & {61.8} &\\
& Solidity & 50.0 & 62.5 & 80.0 & 100.0 & {71.4} &\\

\midrule

\multirow{4}{*}{\shortstack[c]{Ours\\[3pt](Gemini)}}
& Continuity & 87.5 & 90.0 & 61.5 & 83.3 & {78.9} & \multirow{4}{*}{\textbf{77.4}}  \\
& Immutability & 100.0 & 93.3 & 100.0 & 83.3 & {94.7} &\\
& Permanence & 50.0 & 70.0 & 65.4 & 83.3 & {68.4} &\\
& Solidity & 33.3 & 66.7 & 60.0 & 100.0 & {64.3} &\\
\midrule

\multirow{4}{*}{\shortstack[c]{Ours\\[3pt](GPT)}}
& Continuity & 87.5 & 73.3 & 84.6 & 58.3 & 76.3 & \multirow{4}{*}{\textbf{70.3}}  \\
& Immutability & 87.5 & 83.3 & 68.0 & 75.0 & 77.3 \\
& Permanence & 50.0 & 60.0 & 69.2 & 83.3 & 65.8 \\
& Solidity & 50.0 & 45.8 & 65.0 & 100.0 & 58.9 \\

\midrule[0.8pt]



\multirow{4}{*}{VoE}
& Continuity & 50.0 & 53.3 & 61.5 & 50.0 & 55.3 & \multirow{4}{*}{44.9}  \\
& Immutability & 50.0 & 36.7 & 36.0 & 41.7& 38.7 &\\
& Permanence & 25.0 & 46.7 & 38.5 & 50.0 & 42.1 &\\
& Solidity & 50.0 & 41.7 & 50.0 & 16.7 & 42.9 &\\
\midrule


\multirow{4}{*}{\shortstack[c]{Window\\-based\\[3pt]MLLM}}
& Continuity & 75.0 & 56.7 & 34.6 & 16.7 & 44.7 & \multirow{4}{*}{38.9}  \\
& Immutability & 37.5 & 40.0 & 32.0 & 33.3 & 36.0 &\\
& Permanence & 25.0 & 36.7 & 30.8 & 0.0 & 27.6 &\\
& Solidity & 33.3 & 62.5 & 50.0 & 16.7 & 50.0 &\\
\midrule

\multirow{4}{*}{\shortstack[c]{Window\\-based\\MLLM\\(w/ history)}}
& Continuity & 50.0 & 66.7 & 53.8 & 25.0 & 53.9 & \multirow{4}{*}{60.4}\\
& Immutability & 75.0 & 66.7 & 80.0 & 50.0& 69.3 &\\
& Permanence & 50.0 & 70.0 & 65.4 & 33.3 & 60.5 &\\
& Solidity & 33.3 & 54.2 & 70.0 & 50.0 & 57.1 &\\

\bottomrule
\end{tabular}
\vspace{-10pt}
\end{table}

\clearpage
\newpage
\section{Effects of initial noise on reference-guided reconstructions}
\label{app:reconstruction}

Figure~\ref{fig:model_comparison} provides a model-wise breakdown of the reference-guided reconstruction results discussed in Sec.~\mainref{sec:5.5}{5.5} across initial noise levels.
Across all six video generative models, both PhysTrace and human evaluation show decreasing physically impossible classification rates as $\sigma_{\mathrm{init}}$ increases.
Differences across models also become smaller at higher noise levels, further illustrating the broadly consistent trends between PhysTrace and human evaluation.

\begin{figure}[h]
    \centering


    \begin{subfigure}[t]{0.48\columnwidth}
        \centering
        \includegraphics[width=\linewidth]{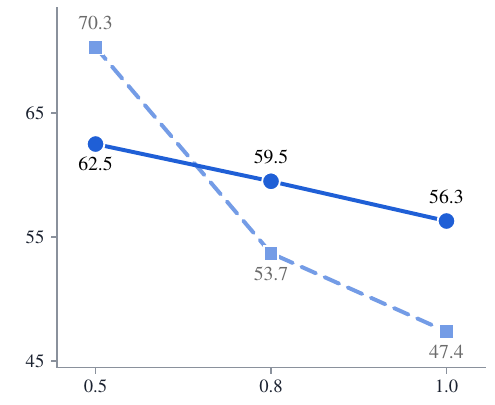}
        \caption{ModelScope-1.7B}
        \label{fig:modelscope}
    \end{subfigure}
    \hfill
    \begin{subfigure}[t]{0.48\columnwidth}
        \centering
        \includegraphics[width=\linewidth]{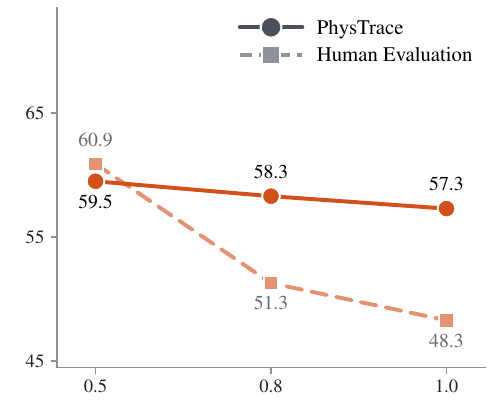}
        \caption{VideoCrafter2}
        \label{fig:videocrafter2}
    \end{subfigure}

    \par\vspace{3pt}

    \begin{subfigure}[t]{0.48\columnwidth}
        \centering
        \includegraphics[width=\linewidth]{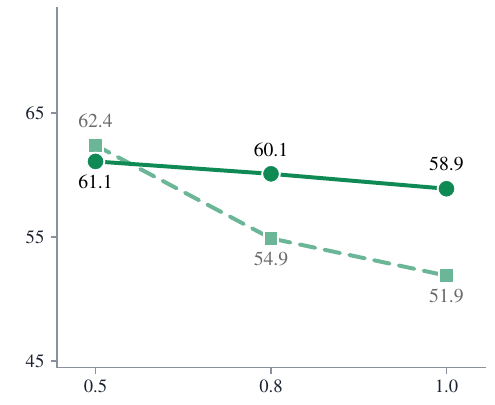}
        \caption{CogVideoX-5B}
        \label{fig:cogvideox}
    \end{subfigure}
    \hfill
    \begin{subfigure}[t]{0.48\columnwidth}
        \centering
        \includegraphics[width=\linewidth]{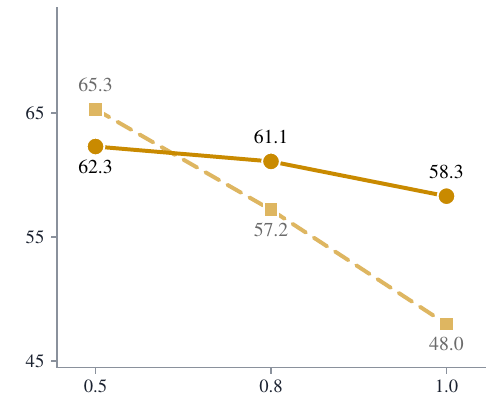}
        \caption{Wan2.1-T2V-1.3B}
        \label{fig:wan}
    \end{subfigure}

    \par\vspace{3pt}

    \begin{subfigure}[t]{0.48\columnwidth}
        \centering
        \includegraphics[width=\linewidth]{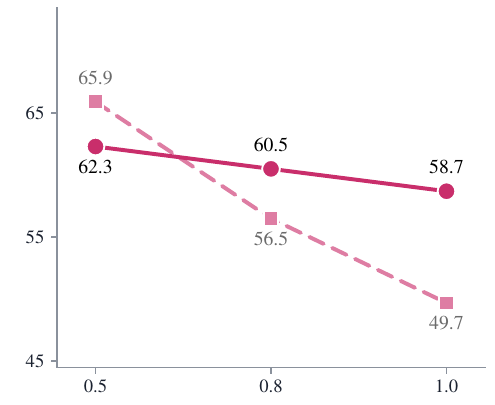}
        \caption{LongCat-Video}
        \label{fig:longcat}
    \end{subfigure}
    \hfill
    \begin{subfigure}[t]{0.48\columnwidth}
        \centering
        \includegraphics[width=\linewidth]{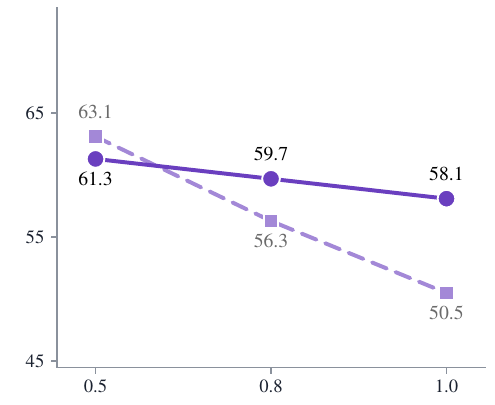}
        \caption{Cosmos-Predict2-2B}
        \label{fig:cosmos}
    \end{subfigure}

    \caption{
        \textbf{Impossible classification rates across models.}
        PhysTrace and human evaluation results across initial noise levels
        ($\sigma_{\mathrm{init}}$) on reference-guided reconstructions
        from the full \textit{IntPhys2-Timing} dataset.
        The x- and y-axes indicate the initial noise level
        ($\sigma_{\mathrm{init}}$) and physically impossible
        classification rate (\%), respectively.
    }
    \label{fig:model_comparison}

\end{figure}

\begin{figure*}[t]
    \centering
    \includegraphics[width=0.95\textwidth]{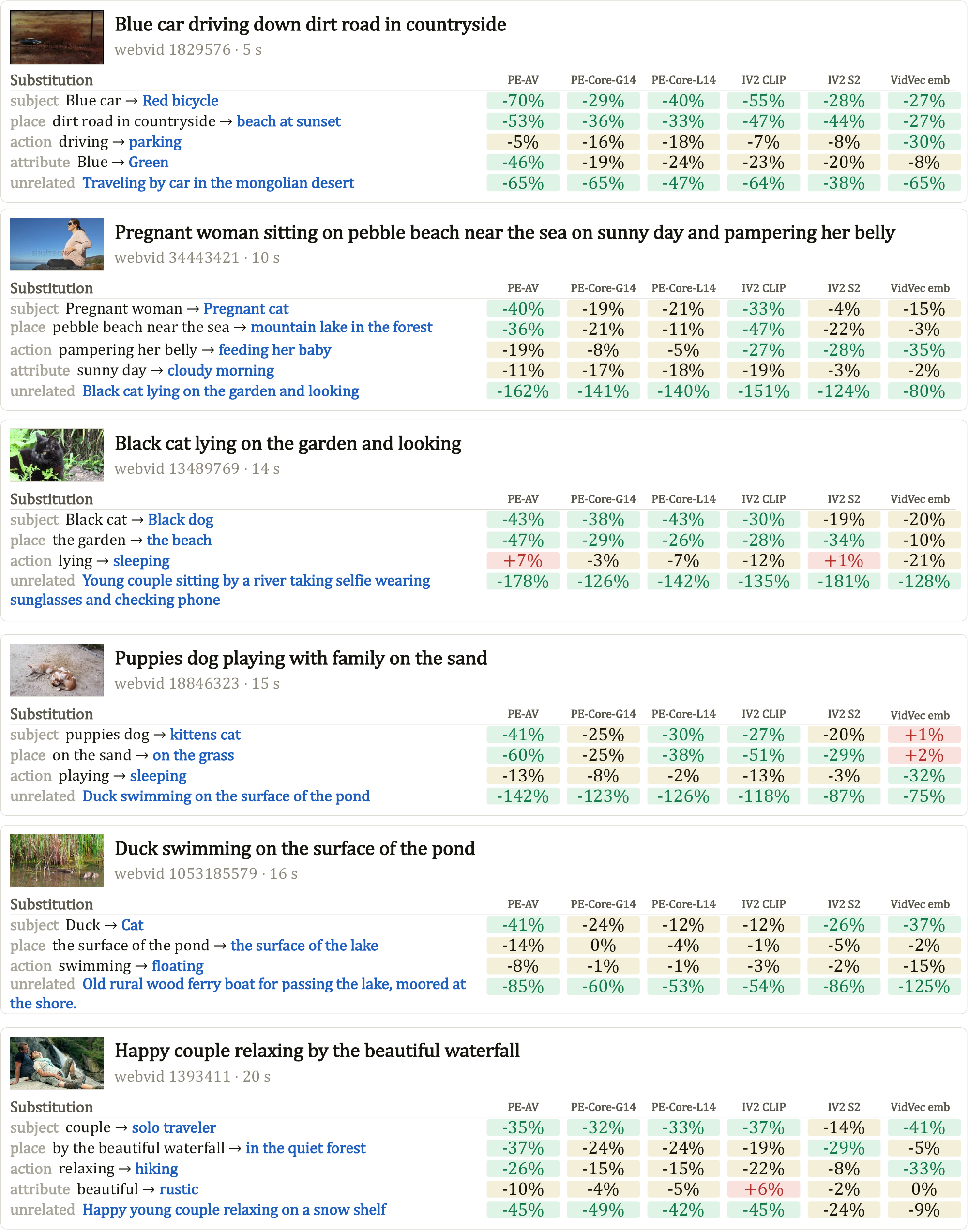}
    \caption{\textbf{Per-clip examples of caption substitution sensitivity on WebVid.}
Each row replaces one caption element (subject, place, action, or attribute); the last row replaces the entire caption with an unrelated sentence. Columns denote alignment scoring models. Cells report $\Delta Det.$, the signed score change relative to the original caption, normalized by the mean drop from unrelated-sentence replacements. \textcolor{mygreen}{Green}: $\Delta Det. \le -25\,\%$; \textcolor{myyellow}{Yellow}: $-25\,\% < \Delta Det. < 0$; \textcolor{myred}{Red}: $\Delta Det. \ge 0$.}
    \label{fig:app1}
\end{figure*}

\begin{figure*}[p]
    \centering

    \begin{subfigure}{\textwidth}
        \centering
        \includegraphics[
            width=\linewidth,
            height=0.49\textheight,
            keepaspectratio
        ]{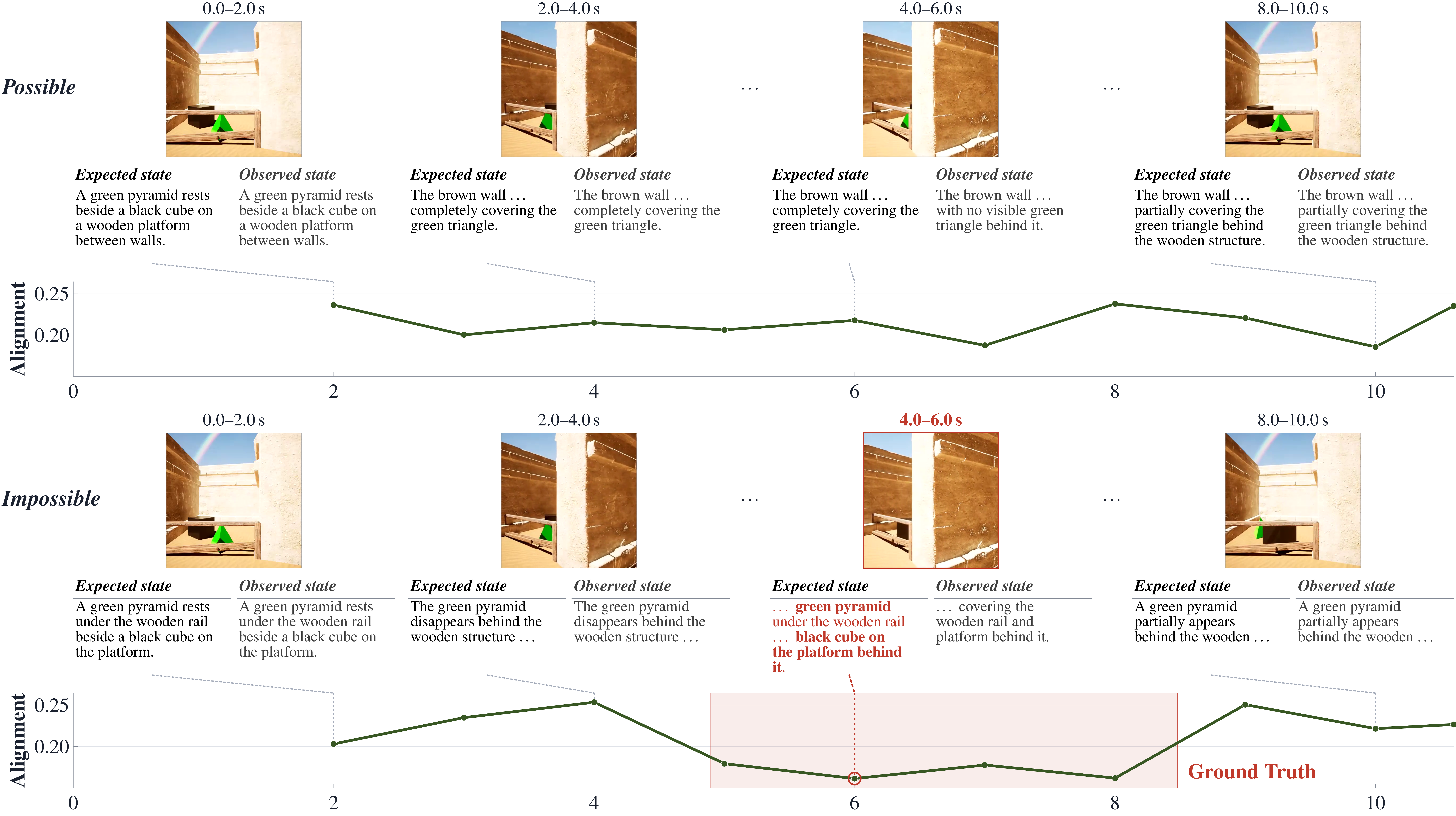}
        \caption{Continuity}
        \label{fig:app2a}
    \end{subfigure}

    \vspace{20pt}

    \begin{subfigure}{\textwidth}
        \centering
        \includegraphics[
            width=\linewidth,
            height=0.49\textheight,
            keepaspectratio
        ]{phystrace_pics/figApp2/immutability.pdf}
        \caption{Immutability}
        \label{fig:app2b}
    \end{subfigure}
    \vspace{0pt}
    \captionsetup{skip=3pt}
    \caption{\textbf{Qualitative examples of history-conditioned physical consistency traces.} Possible/Impossible pairs from IntPhys2-Timing across physical attributes; the x-axis denotes time in seconds.}
    \label{fig:app2}
\end{figure*}
\clearpage

\begin{figure*}[p]
    \ContinuedFloat
    \centering

    \begin{subfigure}{\textwidth}
        \centering
        \includegraphics[
            width=\linewidth,
            height=0.49\textheight,
            keepaspectratio
        ]{phystrace_pics/figApp2/permanence.pdf}
        \caption{Permanence}
        \label{fig:app2c}
    \end{subfigure}

    \vspace{20pt}

    \begin{subfigure}{\textwidth}
        \centering
        \includegraphics[
            width=\linewidth,
            height=0.49\textheight,
            keepaspectratio
        ]{phystrace_pics/figApp2/solidity.pdf}
        \caption{Solidity}
        \label{fig:app2d}
    \end{subfigure}
    \vspace{0pt}
    \captionsetup{skip=3pt}
    \caption{\textbf{Qualitative examples of history-conditioned physical consistency traces.} Possible/Impossible pairs from IntPhys2-Timing across physical attributes; the x-axis denotes time in seconds. (continued)}
\end{figure*}


}{}%

\end{document}